%% file: main.tex
\documentclass{article}
\usepackage[nonatbib, final]{neurips}
\usepackage[numbers]{natbib}

\makeatletter
\renewcommand{\@noticestring}{
  \centering
}
\makeatother

\input{extra_pkgs}

\newif \ifhq
\hqfalse

\newcommand{\nameofmethod}{HunyuanVideo 1.5}
\newcolumntype{C}{>{\centering\arraybackslash}p{1.1cm}}

\newcommand{\secref}[1]{Section~\ref{#1}}

\title{\nameofmethod{} Technical Report}

\author{Tencent Hunyuan Foundation Model Team}

\begin{document}

\maketitle

\input{abs/abs}

\input{intro/intro}

\input{data/data}

\input{model/model}

\input{infra/infra}

\input{training/training}

\input{performance/performance}

\input{inference/inference}

\input{conclusion/conclusion}

\input{contributers/contributer}

\clearpage
{
\bibliographystyle{unsrtnat}
\bibliography{main}
}

\end{document}

%% file: extra_pkgs.tex
\usepackage[export]{adjustbox}
\usepackage[ruled]{algorithm2e}
\usepackage[inline, shortlabels]{enumitem}
\usepackage[T1]{fontenc}
\usepackage{hyperref}
\usepackage{microtype}
\usepackage{pifont}
\usepackage{xcolor}
\usepackage{xurl}
\usepackage{float}
\usepackage{graphicx}
\usepackage{booktabs}
\usepackage{tabularray}
\usepackage{makecell}
\usepackage{array}
\usepackage{rotating}
\usepackage{multicol}
\usepackage{multirow}
\usepackage{listings}
\usepackage{amsmath, amsfonts}
\usepackage{nicefrac}
\usepackage{subcaption}
\usepackage{algorithmic}

\UseTblrLibrary{booktabs}

\makeatletter
\newcommand{\ssymbol}[1]{\@fnsymbol{#1}}
\newcommand{\romanNumeral}[1]{\expandafter\@slowromancap\romannumeral #1@}
\makeatother

%% file: abs/abs.tex

\begin{abstract}

We present \nameofmethod{}, a lightweight yet powerful open-source video generation model that achieves state-of-the-art visual quality and motion coherence with only 8.3 billion parameters, enabling efficient inference on consumer-grade GPUs. 
This achievement is built upon several key components, including meticulous data curation, an advanced DiT architecture featuring selective and sliding tile attention (SSTA), enhanced bilingual understanding through glyph-aware text encoding, progressive pre-training and post-training, and an efficient video super-resolution network. 
Leveraging these designs, we developed a unified framework capable of high-quality text-to-video and image-to-video generation across multiple durations and resolutions.
Extensive experiments demonstrate that this compact and proficient model establishes a new state-of-the-art among open-source video generation models. 
By releasing the code and model weights, we provide the community with a high-performance foundation that lowers the barrier to video creation and research, making advanced video generation accessible to a broader audience. 
All open-source assets are publicly available at \url{https://github.com/Tencent-Hunyuan/HunyuanVideo-1.5}.
\end{abstract}

%% file: intro/intro.tex

%
%

\section{Introduction}

Recent years have witnessed remarkable advancements in video generation models, with closed-source systems like Kling2.5~\cite{kling25}, Veo3.1~\cite{veo31}, and Sora2~\cite{sora2} achieving state-of-the-art (SOTA) performance. In the open-source domain, models such as HunyuanVideo~\cite{hunyuanvideo}, StepVideo \cite{ma2025stepvideot2vtechnicalreportpractice}, and Wan2.2 \cite{wan2025} have emerged as notable contenders. However, despite these strides, most SOTA models remain proprietary, limiting accessibility and community-driven innovation. Wan2.2 \cite{wan2025} employs a hybrid Mixture of Experts (MoE) architecture that leverages two 14-billion-parameter expert models to enhance video quality. Although this MoE architecture enhances visual fidelity by leveraging specialized experts for different denoising stages, it inherently introduces computational inefficiencies, as the model must manage multiple large parameter sets (totaling 27B parameters with 14B activated), leading to significant resource demands. While Wan2.2 \cite{wan2025} has attempted to address this with a more compact 5-billion-parameter variant, which utilizes a high-compression 3D VAE to reduce memory footprint, this lightweight model still exhibits limitations. Specifically, its generative capabilities, particularly in maintaining motion stability across frames and achieving the nuanced aesthetic quality required for professional applications, continue to fall short of practical demands. This situation underscores a critical gap: the absence of an open-source model that successfully balances high-quality output with efficient inference.

To address these challenges, we introduce \nameofmethod{}, an open-source 8.3B video generation model designed to achieve state-of-the-art visual quality while maintaining high inference efficiency. Our framework is built on a two-stage pipeline for high-fidelity video synthesis. The first stage employs a Unified Diffusion Transformer (DiT) that supports both text-to-video and image-to-video generation, producing initial video sequences with resolutions from 480p to 720p and durations of 5 to 10 seconds. The second stage utilizes a dedicated Video Super-Resolution Network to upscale the outputs to 1080p, significantly enhancing the final visual fidelity. The main contributions of this report are summarized as follows:
\begin{itemize}
    \item \textbf{Lightweight High-Performance Architecture:} We propose an efficient architecture that integrates an 8.3B-parameter Diffusion Transformer (DiT) with a 3D causal VAE, achieving compression ratios of 16× in spatial dimensions and 4× along the temporal axis. This design minimizes parameter count while fully leveraging the potential of the model, yielding performance comparable to state-of-the-art video generation models.
    
    \item \textbf{Video Super-Resolution Enhancement:} We develop an efficient few-step super-resolution network that upscales outputs to 1080p. It enhances sharpness while correcting distortions, thereby refining details and overall visual texture.
    \item \textbf{Sparse Attention Optimization:} We introduce a novel Selective and Sliding Tile Attention (SSTA) mechanism that dynamically prunes redundant spatiotemporal tokens. This significantly reduces computational overhead for long video sequences and accelerates inference, achieving an end-to-end speedup of $1.87 \times $ in 10-second 720p video synthesis compared to FlashAttention-3 \cite{shah2024flashattention3fastaccurateattention}.

    \item \textbf{Enhanced Multimodal Understanding:} Our framework utilizes a large multimodal model for precise bilingual (Chinese-English) understanding, combined with ByT5 for dedicated glyph encoding to enhance text generation accuracy in videos. Additionally, detailed bilingual captions are generated for both images and videos.
    
    \item \textbf{End-to-End Training Optimization:} We demonstrate that the Muon optimizer \cite{kimiteam2025kimik2openagentic} significantly accelerates convergence in video generation model training, while the multi-phase progressive training strategy—spanning from pre-training to post-training stages—enhances motion coherence, aesthetic quality, and alignment with human preferences, thereby enabling the production of professional-grade content.
    
\end{itemize}

This report presents a comprehensive pipeline, beginning with meticulous data preparation—including curation, filtering, and captioning, as detailed in \textbf{\secref{sec:data}}. \textbf{\secref{sec:model}} introduces the core architecture and key algorithms, such as the unified diffusion transformer, the video super-resolution network, and the mechanisms of sparse attention.  \textbf{\secref{sec:training}} describes training strategies for text-to-video and image-to-video tasks, covering both pre-training and post-training methodologies. Finally, \textbf{\secref{sec:performance}} provides a rigorous evaluation of the model’s performance against state-of-the-art models using quantitative metrics and qualitative benchmarks.

%% file: data/data.tex
\section{Data Preparation}\label{sec:data}

\subsection{Data Acquisition and Filtering}

Our training dataset comprises both image and video data. For image data, we adopted the acquisition and processing pipeline outlined in~\cite{cao2025hunyuanimage}, curating 5 billion images from a pool of over 10 billion for pre-training, with 1 billion reserved for subsequent stages. 
For video data, building upon the pipeline in~\cite{hunyuanvideo}, we significantly expanded the data volume while refining the filtering mechanisms to enhance quality. 
The specifics of these processes are detailed below.




\textbf{Data Acquisition.}
 To continuously improve the model's performance, we prioritized diversity and quality during data acquisition. We sourced raw videos from a variety of channels, ensuring comprehensive coverage across diverse content, filming techniques, camera movements, styles, and scenarios. After basic deduplication and the removal of corrupted files, we obtained more than 10 million hours of raw video data.

To address the high variance in raw video lengths and optimize training efficiency, we segmented the videos into consistent clips ranging from 2 to 10 seconds. Specifically, we utilized PySceneDetect~\cite{PySceneDetect} combined with a custom segmentation operator to detect scene boundaries. To eliminate any remaining clips containing transition effects, we applied a secondary filtering stage using a specialized transition classifier.

To mitigate artifacts such as subtitles, logos, and watermarks, we applied spatial cropping to exclude affected regions. To maintain compositional integrity, we discarded clips where the cropped area retained less than 60\% of the original frame.

\textbf{Data Curation.}
To guarantee the high quality of our training data, we implemented a rigorous, multi-stage filtering pipeline. This pipeline consists of three distinct levels:
\begin{itemize}
    \item \textbf{Basic Filtering}: We employed specialized filters to remove videos with padding borders, stitching artifacts, grid layouts (collages), and static or low-motion scenes.
    \item \textbf{Visual Quality Filtering}: We developed a comprehensive quality assessment operator that evaluates videos across four dimensions: sharpness, detail retention, noise \& artifacts, and dynamic range. This allows us to discard clips with poor visual quality.
    \item  \textbf{Aesthetic Filtering}: We applied an aesthetic scoring operator based on~\cite{dover} to evaluate the aesthetic quality of the videos, filtering out those with low aesthetic scores.
\end{itemize}
Upon completion of these filtering stages, approximately 800 million high-quality video segments remained for pre-training.

\subsection{Captioning}

High-quality captions play a crucial role in video generation, significantly influencing precise instruction following and generation quality. To generate precise and comprehensive descriptions across diverse data modalities, we developed three specialized captioning models, each tailored for a specific function:
\begin{itemize}
    \item \textbf{Image Captioning}: Generates descriptions for static images, leveraging the same methodology in HunyuanImage-3.0 \cite{cao2025hunyuanimage}. 
    \item \textbf{Video Captioning}: Produces a highly structured, multi-component description. These encompass multi-level textual narratives alongside a detailed set of cinematic and aesthetic properties (e.g., shot type, shot angle, composition, lighting, style, color palette, atmosphere).
    \item \textbf{Image-to-Video (I2V) Instructional Captioning}: This novel module generates text describing the temporal evolution or transformation from the initial frame, detailing changes in both foreground subjects and the background environment.
\end{itemize}

\begin{figure}[h]
    \centering
    \includegraphics[width=\linewidth]{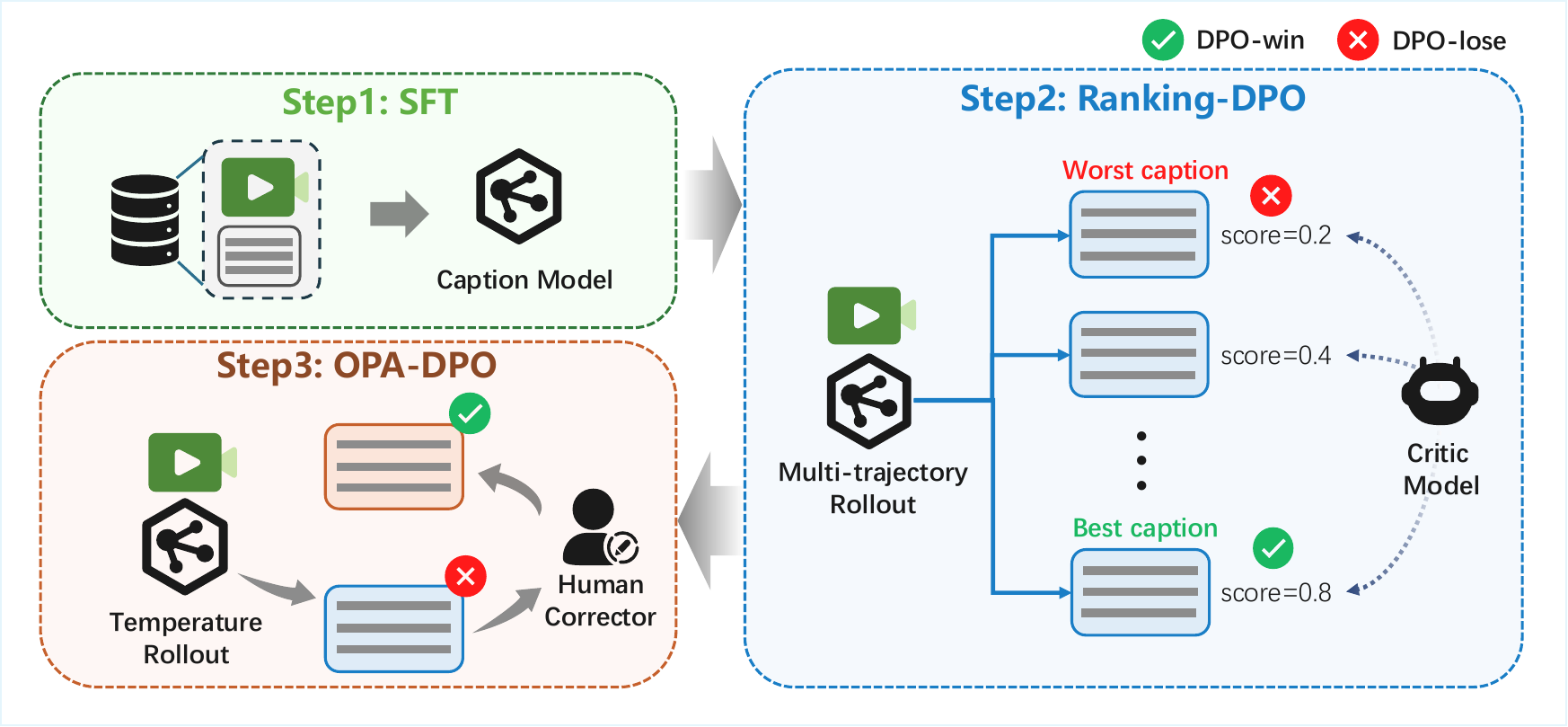}
    \caption{ Caption Model Post-training Pipeline.}
    \label{fig:caption}
\end{figure}

\textbf{Addressing the Richness-Hallucination Trade-off.} 
A significant and persistent challenge in this domain is the inherent trade-off between descriptive richness and factual accuracy. Specifically, attempts to generate greater detail often increase the propensity for hallucination or factual inconsistencies. To overcome this, we integrate Reinforcement Learning (RL), specifically OPA-DPO \cite{yang2025mitigating}, into our caption model post-training pipeline, as shown in Figure \ref{fig:caption}. This approach is designed to strike an optimal balance, maximizing descriptive detail while simultaneously minimizing hallucinations and factual errors.


\textbf{Camera Movement Description.} We developed a specialized recognition model to analyze camera dynamics. This model is capable of identifying multiple camera movement types, operating at both a clip-level (for the whole video) and a sequential (over time) level. High-confidence predictions are converted into natural language descriptions and subsequently integrated into the structured video captions. The objective of this integration is to empower generative models with controllable camera movement.





%% file: model/model.tex
\section{Model Design}\label{sec:model}

We propose a two-stage framework for high-fidelity video synthesis. In the first stage, we employ a diffusion transformer (DiT) model with 8.3 billion parameters, designed for multi-task learning. The architectural hyperparameters of \nameofmethod{} are detailed in Table \ref{tab:model_settings}. Subsequently, a video super-resolution network is utilized to further enhance visual quality. Comprehensive elaborations are provided in the following subsection.

\begin{table}[t]
  \centering
  \footnotesize
  \caption{Architecture hyperparameters for the \nameofmethod{}.}
  \begin{tabular}{ccccc}
  \toprule
  \textbf{\makecell{Dual-stream Blocks}}  & \textbf{\makecell{Model Dimension}} & \textbf{\makecell{FFN Dimension}} & \textbf{\makecell{Attention Heads}} & \textbf{\makecell{Head dim}}	\\
  \midrule
  54 & 2048 & 8192 & 16 & 128 \\
  \bottomrule
  \end{tabular}%
  \label{tab:model_settings}
\end{table}

\begin{figure}[h]
    \centering
    \includegraphics[width=\linewidth]{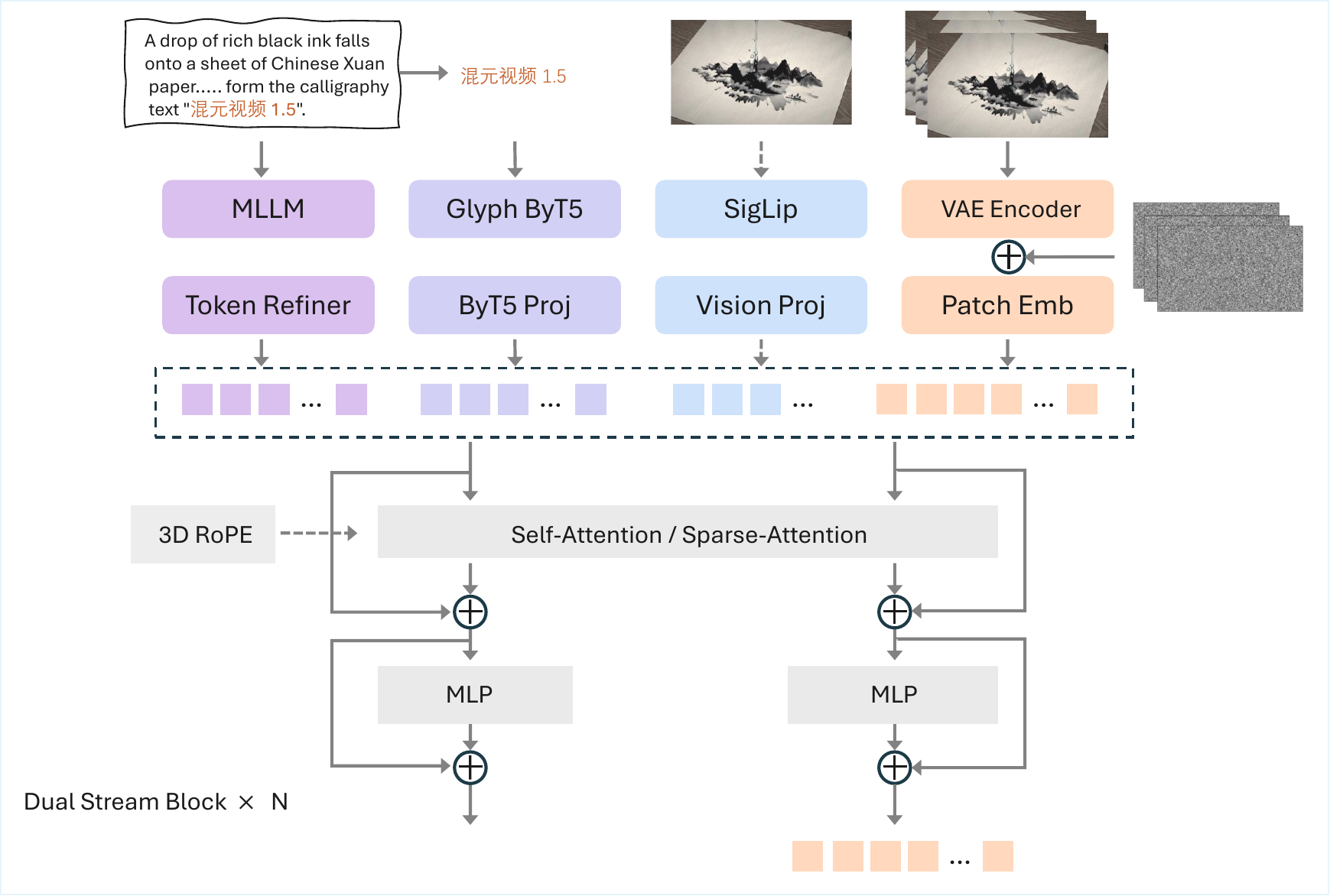}
    \caption{Architecture of the Unified Diffusion Transformer.}
    \label{fig:dit}
\end{figure}

%

\subsection{Unified Diffusion Transformer}

\textbf{Multi-Task Training.}  We developed a unified architecture for jointly training text-to-image (T2I), text-to-video (T2V), and image-to-video (I2V) tasks. For the I2V task, the reference image is integrated into the model via two complementary strategies: (1) VAE-based encoding, where the image latent is concatenated with the noisy latent along the channel dimension to leverage its exceptional detail reconstruction capacity; and (2) SigLip-based feature extraction, where semantic embeddings are concatenated sequentially to enhance semantic alignment and strengthen instruction adherence in I2V generation. A learnable type embedding is introduced to explicitly distinguish between different types of conditions.

\textbf{VAE.} We introduce a causal 3D transformer architecture designed for joint image-video encoding, which achieves a spatial compression ratio of 16$\times$ and a temporal compression ratio of 4$\times$, with a latent channel dimension of 32.

\textbf{Text encoder.} The model employs a dual-channel text encoder, combining a general textual encoder. This design leverages a visual-language multimodal encoder Qwen2.5-VL \cite{bai2025qwen25vltechnicalreport} to achieve a deeper understanding of scene descriptions, character actions, and specific requirements. Additionally, the incorporation of a multilingual Glyph-ByT5 \cite{liu2024glyphbyt5customizedtextencoder} text encoder further strengthens the model’s ability to render text accurately across various languages. This synergistic design enables the model to learn from both high-level semantic representations and fine-grained language-specific features, leading to a more comprehensive and robust text comprehension capability.

\textbf{Sparse Attention.} The computational complexity of standard self-attention scales quadratically with sequence length, posing significant efficiency bottlenecks for transformer-based video generation models. Given the inherent spatiotemporal redundancy in video data, replacing full attention with sparse attention presents a well-motivated optimization strategy. To synergistically combine the benefits of static local window priors and dynamic global adaptive selection, we propose a novel sparse attention method termed SSTA (Selective and Sliding Tile Attention). SSTA features a parameter-free design, enabling seamless integration at any training stage. In our implementation, sparse training is incorporated during the distillation phase, which more effectively preserves output quality while maintaining high computational efficiency. The SSTA algorithm comprises four key steps, as described in Algorithm ~\ref{alg:ssta}: 3D Block Partition, Selective Mask Generation, STA \cite{zhang2025fastvideogenerationsliding} Mask Generation and Block-Sparse Attention. We propose an engineered acceleration toolkit \cite{flex_block_attn2025} for sparse attention mechanisms, utilizing the ThunderKittens \cite{spector2024thunderkittenssimplefastadorable} framework to efficiently implement the flex\_block\_attention algorithm.

\begin{algorithm}[tb]
\caption{Selective and Sliding Tile Attention}
\label{alg:ssta}
\begin{algorithmic}[1]
\REQUIRE 
Query/Key/Value tensor $Q, $K, $V \in \mathbb{R}^{h \times F \times H \times W \times D}$, \\
Block size $N = tile_t \times tile_h \times tile_w$, \\
Block num $B = (F/tile_t) \times (H/tile_h) \times (W/tile_w)$, \\
Window size $WS = (w_t, w_h, w_w)$, \\
Top-$k$.

\vspace{0.5em}
\STATE \textbf{(1) 3D Block Partition}
    \STATE Partition $Q$ and $K$ into blocks of size $N$
    \STATE $Q_b, K_b \gets \text{split\_blocks}(Q, K, N)$

\vspace{0.5em}
\STATE \textbf{(2) Selective Mask Generation}
    \STATE $\bar{Q}_b, \bar{K}_b \gets \text{adaptive\_pool}(Q_b), \text{adaptive\_pool}(K_b)$
    \STATE $\mathrm{Score}_s = \bar{Q}_b \bar{K}_b^\top \in \mathbb{R}^{h \times B \times B}$  \COMMENT{Q-K block-wise similarity}
    \STATE $\mathrm{Score}_r = \frac{1}{N-1} \sum_{\substack{ i=1 \\ j \neq i}}^N [\bar{K}_b \bar{K}_b^\top]_{ij} \in \mathbb{R}^{h \times 1 \times B}$  \COMMENT{K-K block-wise redundancy}
    \STATE $\mathrm{Score}_i = \lambda \cdot \mathrm{Score}_s - \beta \cdot \mathrm{Score}_r$  \COMMENT{Block importance}
    \STATE $\mathrm{Selected\_indexes} \gets \text{Topk}(\mathrm{Score_i}, k)$
    \STATE $M_{sel} \gets \text{Indexes2mask}(\mathrm{Selected\_indexes})$

\vspace{0.5em}
\STATE \textbf{(3) STA (Sliding Tile Attention) Mask Generation}
    \FOR{each block pair $(i, j)$}
        \IF{$j$ within local window of $i$ defined by $WS=(w_t, w_h, w_w)$}
            \STATE $M_{\mathrm{sta}}[i, j] \gets 1$
        \ELSE
            \STATE $M_{\mathrm{sta}}[i, j] \gets 0$
        \ENDIF
    \ENDFOR

\vspace{0.5em}
\STATE \textbf{(4) Combine Masks \& Perform Block-Sparse Attention}
    \STATE $M_{combined} \gets M_{sel} \land M_{sta}$ \COMMENT{Combine $M_{sel} \in \mathbb{R}^{h \times B \times B}$ and $M_{sta} \in \mathbb{R}^{B \times B}$}
    \STATE $O \gets \text{flex\_block\_attention}(Q, K, V, M_{combined})$

\vspace{0.5em}
\RETURN $O$

\end{algorithmic}
\end{algorithm}

\textbf{Optimizer.} The Muon optimizer \cite{kimiteam2025kimik2openagentic} is used in this work to achieve faster convergence. We observe that it attains a lower training loss than AdamW in only half the number of training steps, while also yielding superior performance across multiple text-to-image benchmarks. To ensure training stability, a weight decay of 0.01 is applied.

\begin{figure}[t]
    \centering
    \includegraphics[width=\linewidth]{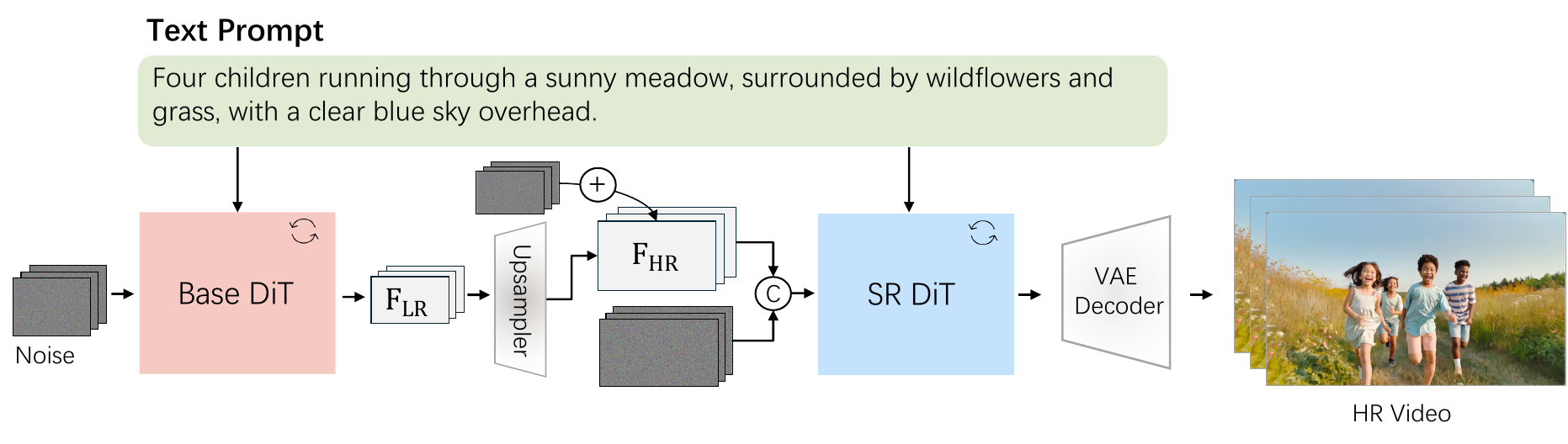}
    \caption{The pipeline of a cascaded video super-resolution model.}
    \label{fig:vsr}
\end{figure}

\subsection{Video Super-Resolution}

To further improve training and inference efficiency while meeting the demand for high-resolution video generation, we introduce a cascaded Video Super-Resolution (VSR) model that enhances fine-grained visual details and textures in videos generated by text-to-video and image-to-video based models. We follow the architecture of the main model—an 8.3B Video Diffusion Transformer—and perform super-resolution in the latent space. We inject the LR latents using a channel concatenation approach. To spatially align the LR and HR latents, we separately train a latent upsample block. The final high-quality 1080p video is obtained by decoding these HR latents through the VAE decoder, as shown in Figure~\ref{fig:vsr}.

%% file: infra/infra.tex

%% file: training/training.tex
\section{Model Training}\label{sec:training}

\subsection{Pre-training}

We establish a robust and versatile foundation for HunyuanVideo 1.5 through a structured pre-training strategy that systematically develops core generative capabilities across multiple modalities. This phase employs a progressive, multi-stage approach encompassing text-to-image (T2I), text-to-video (T2V), and image-to-video (I2V) tasks. By scaling spatial resolution, temporal length, and frame rate progressively and leveraging mixed-task training with carefully balanced data ratios, we equip the model with strong semantic alignment, visual diversity, and temporal coherence, forming a solid base for subsequent specialization and refinement.

\textbf{T2I.} We begin by training the T2I task, which includes two resolution stages (Stage I \& II in Table \ref{tab:training_stages}): 256p and 512p. For each stage, we utilize training images with various aspect ratios and organize them into buckets to ensure efficient data sampling. The T2I task enables the model to learn robust semantic alignment between text and images, which we find significantly accelerates the convergence and performance of subsequent T2V \& I2V training.

\textbf{T2V \& I2V.} Throughout the pretraining phase, we employ a mixed training regimen combining T2I, T2V, and I2V tasks, with a task ratio of 1:6:3. This configuration leverages large-scale T2I data to enhance semantic understanding and generative diversity, while ensuring sufficient video-specific modeling through T2V and I2V tasks. We adopt a progressive multi-stage strategy (Stages III to VI in Table \ref{tab:training_stages}), starting from 256p resolution at 16 fps and progressively advancing to 480p and 720p at 24 fps, with video durations ranging from 2 to 10 seconds. This structured scaling of spatiotemporal resolution facilitates stable convergence and strengthens the model’s capacity for detailed, coherent video synthesis. Similar to prior studies \cite{esser2024scaling}, we observe that flow matching-based training is particularly sensitive to the shift hyper-parameter when video token lengths vary across stages. To maintain training stability under such varying sequence conditions, we carefully design a series of shift scheduling strategies that adapt to different token lengths during the progressive training process.

\begin{table}[t]
\centering
\caption{The training process for the three tasks: T2I, T2V, I2V. In the "Data volume" column, the number before the "/" indicates the amount of video data, and the number after the "/" indicates the amount of image data. "M" stands for million.} 
\vspace{3pt}
\label{tab:training_stages}
\begingroup
\setlength{\tabcolsep}{13pt}
\begin{tabular}{l|cccc}
    \toprule
     Stage & Stage & Training Resolution & Data Volumn & Task  \\
    \midrule
     I  & Pretrain  & 256p & 5 billion & T2I \\
     II & Pretrain  & 512p & 1 billion & T2I \\
     \midrule
     III & Pretrain & 256p 16fps 2$\sim$10s & 800M & T2V/I2V/T2I \\ 
     IV &  Pretrain & 480p 16fps 2$\sim$10s & 200M & T2V/I2V/T2I \\
     V &  Pretrain & 720p 16fps 2$\sim$10s & 100M & T2V/I2V/T2I \\
     VI &  Pretrain & 720p 24fps 2$\sim$10s & 100M & T2V/I2V/T2I \\
     \midrule
     VII &  CT & 480p/720p 24fps 2$\sim$10s & 1M & T2V \\
     VIII &  CT & 480p/720p 24fps 2$\sim$10s & 1M & I2V \\
    \bottomrule
    \end{tabular}
\endgroup
\end{table}

\subsection{Post-training}

\begin{figure}[tbp]
    \centering
    \includegraphics[width=\linewidth]{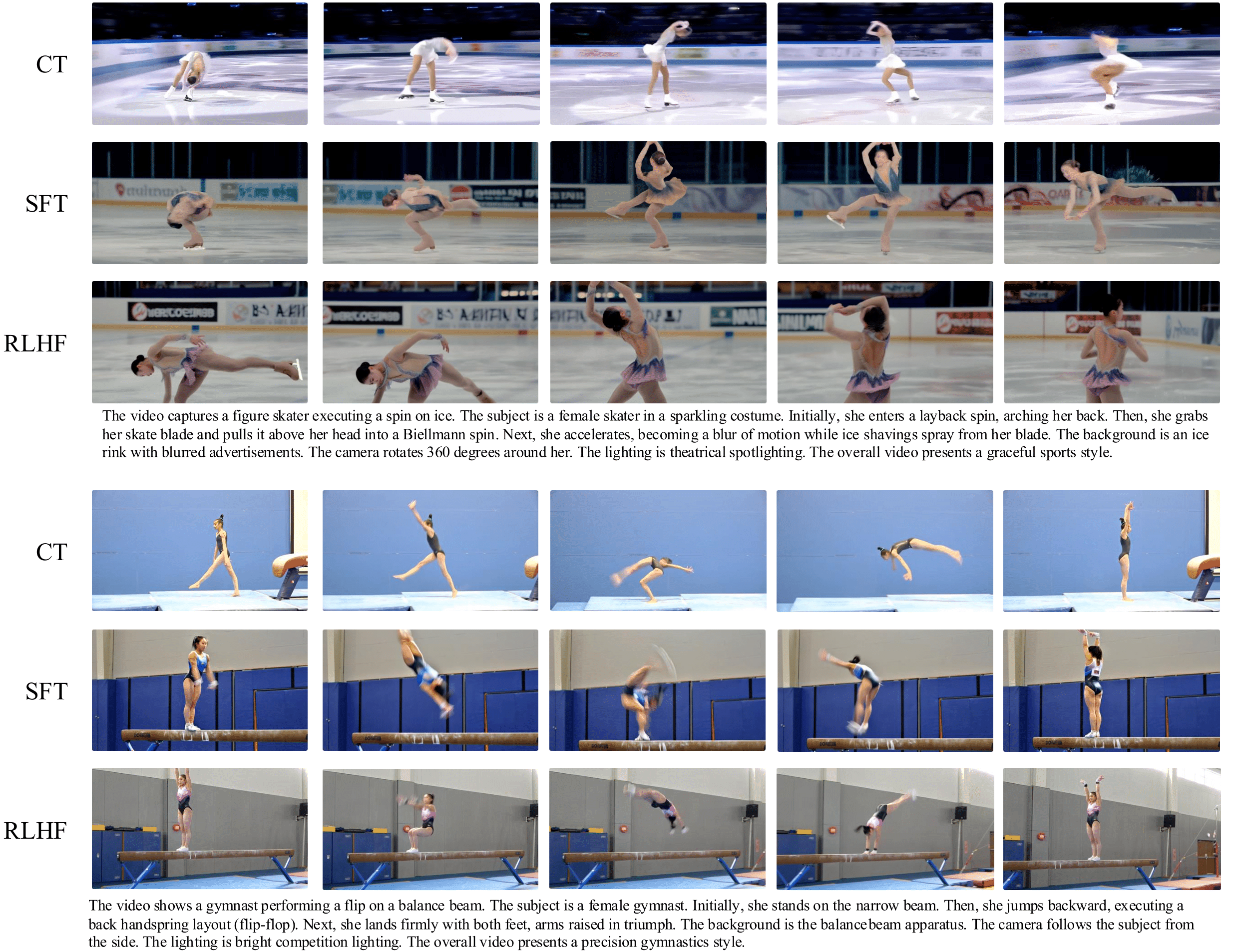}
    \caption{Visualization during different post-training stages}
    \label{fig:posttrain}
\end{figure}

We conduct a comprehensive post-training pipeline to further enhance the generative quality and task-specific performance of HunyuanVideo 1.5. This multi-stage process consists of continuing training (CT), supervised fine-tuning (SFT), and human feedback alignment (RLHF), applied separately for text-to-video (T2V) and image-to-video (I2V) tasks. As shown in Figure \ref{fig:posttrain}, we present the visualization results during different post-training stages. By progressively refining the model distribution through these dedicated stages, we systematically steer the model toward the optimal output distribution for each task, significantly improving motion coherence, visual fidelity, and alignment with human preferences.

\textbf{CT.}
We conduct both CT and SFT stage in 480p and 720p resolution. In the CT stage, we leverage 1 million high-quality video clips per task (as detailed in Table \ref{tab:training_stages}, Stages VII \& VIII), balanced across diverse categories and sourced from premium datasets. For T2V, we further prioritize clips with high dynamic motion to strengthen temporal modeling. For I2V, we introduce instructional captions that focus exclusively on describing motion and transformations compared to the first frame, thereby enhancing action fidelity. Both T2V and I2V CT processes are initialized from the same optimally pre-trained checkpoint, enabling both tasks to build upon a shared high-quality foundation. 

\textbf{SFT.}
During SFT, we utilize rigorously selected clips per task, filtered strictly considering aesthetic appeal, clarity, and motion smoothness. Through CT and SFT, we observe evident improvements in output stability, visual quality, aesthetic appeal, and temporal consistency for both generation tasks. 

\textbf{RLHF.}
We employ Reinforcement Learning from Human Feedback (RLHF) strategies tailored for our I2V and T2V tasks, both mainly aimed at correcting artifacts and enhancing motion quality.

For the I2V task, we apply \textbf{online reinforcement learning (RL)} during post-training to correct structural and motion artifacts. Our approach begins with a curated prompt set spanning 100+ categories, constructed from high-aesthetic images. Candidate prompts are first generated via a vision-language model (VLM), then manually verified to ensure strict text-image consistency. We fine-tune a VLM-based reward model to evaluate videos across four key dimensions: text alignment, image alignment, visual quality, and motion dynamics. During RL training, we employ mixed sampling strategies by varying both random seeds and CFG scales, and adopt a hybrid ODE–SDE solver (MixGRPO~\cite{li2025mixgrpounlockingflowbasedgrpo}) to enrich exploration while maintaining sampling quality. This RLHF process yields consistent improvements across all evaluation metrics, with particularly notable gains in motion realism. 

Building on this, our T2V alignment strategy uses a more comprehensive \textbf{hybrid offline-then-online} approach, designed to address the increased complexity of T2V motion artifacts. We found that existing reward models struggle to effectively differentiate fine-grained motion quality. Therefore, we first conduct an offline optimization stage using Direct Preference Optimization (DPO). For this, we curate a balanced $O(10K)$ prompt set (from LLMs-generated prompts and training video captions) covering diverse dimensions (motion, scene, subject, etc.). Using a selected high-quality SFT checkpoint, we generate N video candidates per prompt to create non-repetitive pairs. These pairs are manually annotated (GSB) for semantic fidelity, motion quality, and aesthetics. Applying Direct Preference Optimization (DPO)~\cite{wallace2024diffusion} to this high-quality paired data significantly reduces motion artifacts and establishes a superior policy starting point. Finally, we conduct an online optimization stage, adopting the exact same online RL framework developed for I2V to further enhance the model's visual quality and semantic-text alignment.

\subsection{Video Super-Resolution.}

We begin by initializing the weights of our video super-resolution model with a pretrained T2V video model to ensure effective training. For dataset construction, a diverse collection comprising 1 million high-quality video clips is curated. These clips span a wide range of scenarios and resolutions from 1K to 4K, each lasting between 3 and 10 seconds at 24 frames per second.  In addition to video data, a set of high-resolution images is incorporated during training to enhance the model’s capability to generate fine visual details.

\begin{figure}[t]
    \centering
    \includegraphics[width=\linewidth]{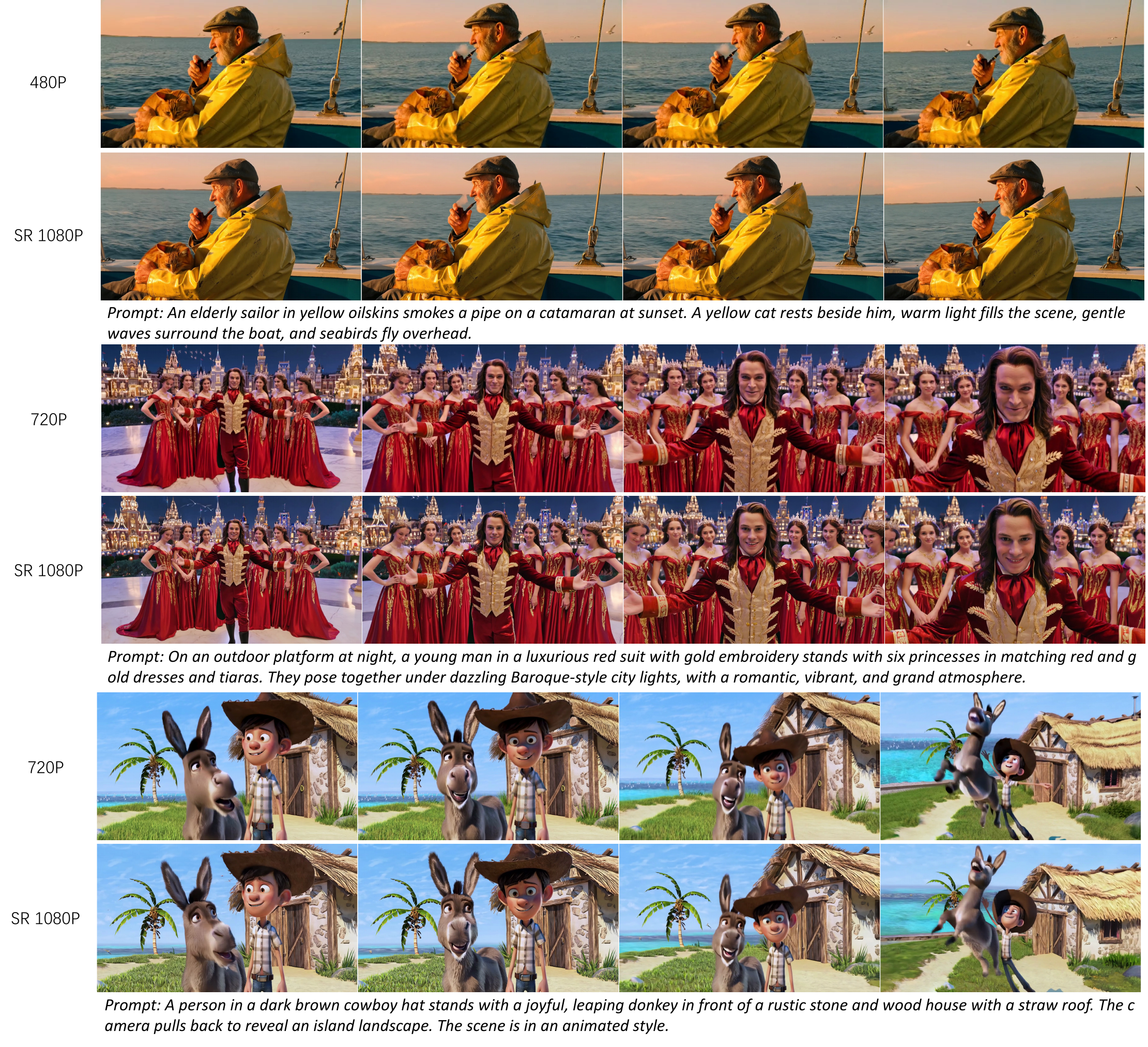}
    \caption{Visualization results of the cascaded video super-resolution model.}
    \label{fig:vsr_case}
\end{figure}

For input preparation, we concatenate the low-resolution latents and noise along the channel dimension as input to the DiT model.  All weights of the video super-resolution model are fully trainable throughout the training process, following the flow matching paradigm.  As shown in Figure~\ref{fig:vsr_case}, our method significantly enhances visual quality, motion stability, and overall temporal coherence.

%% file: performance/performance.tex
\section{Model Performance}\label{sec:performance}

We adopt two evaluation methods: Rating and GSB (Good/Same/Bad). The Rating method provides a comprehensive assessment of the model from multiple perspectives, enabling us to better diagnose its shortcomings. The GSB approach is widely used to evaluate the relative performance of two models based on overall video perception quality.

\subsection{Rating}
As shown in Table \ref{tab:t2v_rating} and Table \ref{tab:i2v_rating}, we assess text-to-video generation using a comprehensive rating methodology that considers five key dimensions: text-video consistency, aesthetic quality of individual frames, visual quality, structural stability, and motion effects. For image-to-video generation, the evaluation encompasses image-video consistency, instruction responsiveness, visual quality, structural stability, and motion effects.

This systematic evaluation framework enables us to quickly diagnose model shortcomings and identify specific areas for improvement. By analyzing performance across these dimensions, we can effectively guide optimization efforts to enhance overall model quality and ensure the generated videos better fulfill task requirements.

\begin{table}[htb]
\centering
\caption{The rating of HunyuanVideo 1.5 T2V model and its competitors} 
\label{tab:t2v_rating}
\begin{tabular}{l|cccccc}
    \toprule
Dimension  & HY1.5 720p & Wan2.2 & Kling2.1 Master & Seedance Pro & Veo3 \\
\midrule
Instruction following & 61.57 & 44.07 & 50.03 & 53.19 & 73.77 \\
Aesthetic quality & 63.30 & 65.98 & 68.00 & 68.22 & 67.98 \\
Visual quality & 57.35 & 56.37 & 59.68 & 60.20 & 58.64 \\
Structural stability & 79.75 & 73.75 & 66.74 & 68.69 & 75.62 \\ 
Motion effects & 57.67 & 53.08 & 58.59 & 55.17 & 60.81 \\
     \bottomrule
    \end{tabular}


\end{table}

\begin{table}[htb]
\centering
\caption{The rating of HunyuanVideo 1.5 I2V model and its competitors. "HY1.5 480pSR" means we first generate 480p video and upscale it to 720p using our video super-resolution model.} 
\label{tab:i2v_rating}
\resizebox{\textwidth}{!}{
    \begin{tabular}{l|cccccc}
    \toprule
Dimension & HY1.5 480pSR & HY1.5 720p & Wan2.2 & Kling2.1 Master & Seedance Pro & Veo3\\
\midrule
Instruction following & 63.11 & 63.05 & 56.19 & 68.43 &  62.90 & 67.86\\
Image consistency & 68.82 & 72.07 & 73.53 & 64.09 & 73.06 & 72.19 \\
Visual quality & 59.87 & 60.33 & 58.31 & 59.28 & 58.95 & 59.29 \\
Structural stability & 70.13 &  66.67 & 69.03 & 59.71 & 68.01 & 69.25\\ 
Motion effects & 57.60 & 58.62 & 57.41 & 57.36 & 60.47 & 60.91 \\
     \bottomrule
    \end{tabular}
}
\end{table}

\subsection{GSB}

In our experiments, we carefully construct 300 diverse text prompts and 300 image samples to cover balanced application scenarios for both text-to-video and image-to-video tasks. For each prompt or image input, an equal number of video samples are generated by each model in a single run to ensure comparability. To maintain fairness, inference is performed only once per input without any cherry-picking of results. All competing models are evaluated using their default configurations. The evaluation is conducted by over 100 professional assessors. The results are shown in Table \ref{tab:t2v_gsb} and \ref{tab:i2v_gsb}.
\begin{table}[htb]
\centering
\caption{GSB of HunyuanVideo 1.5 720P T2V model and its competitors} 
\label{tab:t2v_gsb}
\resizebox{0.7\textwidth}{!}{
    \begin{tabular}{l|cccc}
    \toprule
Compared Model & Wan2.2 & Kling2.1 Master & Seedance Pro & Veo3 \\
\midrule
HY Better & 34.90\% & 31.55\% & 31.67\% & 24.64\% \\
Other better &  17.78\% & 18.95\% & 20.65\% & 34.96\% \\
Equally Bad & 40.70\% & 42.76 \% & 38.61 \% & 30.71\% \\
Equally good & 6.62\% & 6.75\% & 9.06\% & 9.69\% \\
\midrule
HY Win Rate & 17.12\% & 12.6\% & 11.02\% & -10.32\%\\
     \bottomrule
    \end{tabular}
}
\end{table}

\begin{table}[htb]
\centering
\caption{GSB of HunyuanVideo 1.5 720P I2V model and its competitors} 
\label{tab:i2v_gsb}
\resizebox{0.7\textwidth}{!}{
    \begin{tabular}{l|cccc}
    \toprule
Compared Model & Wan2.2 & Kling2.1 Master & Seedance Pro & Veo3 \\
\midrule
HY Better & 45.60\% & 40.60\% & 30.62\% & 37.44\% \\
other better & 32.95\% & 30.88\% & 36.39\% & 41.05\% \\
Equally Bad & 18.60\% & 22.73\%  & 26.61\% & 18.44\%\\
Equally good & 2.85\% & 5.79\% & 6.38\% & 3.07\% \\
\midrule
HY Win Rate & 12.65\% & 9.72\% & -5.77\% & -3.61\% \\
     \bottomrule
    \end{tabular}
}
\end{table}


\section{Inference Speed and GPU Memory Requirements}
\label{sec:inference-speed-and-memory-requirements}


This section presents a comprehensive analysis of the inference speed and memory requirements for \nameofmethod{} under various configurations. 
All speed measurements are conducted on the classifier-free guidance (CFG) distilled model using 8 NVIDIA H800 GPUs with context parallelism enabled across all devices.

\paragraph{Inference Speed without Engineering Acceleration.} 
To better demonstrate the effectiveness of the proposed sparse attention mechanism, we evaluate inference speed without engineering-level acceleration.
No engineering-level acceleration or custom optimization is applied beyond the described model methods and standard distributed inference. All reported results reflect unoptimized, out-of-the-box research code performance.
For configurations without sparse attention, Flash Attention v3~\cite{shah2024flashattention3fastaccurateattention} is adopted as the attention backend.
Inference speed is measured as the wall-clock time per diffusion step, averaged across multiple runs.
Table \ref{tab:inference-time-no-accel} presents the inference speed results under different configurations.
It can be observed that without engineering acceleration, the model already achieves competitive inference speeds.
This is attributed to the compact 8.3B-parameter architecture that is intrinsically efficient and the high VAE compression rate that significantly reduces the number of tokens required for computation.
In addition, sparse attention effectively reduces inference time for long-context sequences, particularly evident in the configurations of 241 frames in Table \ref{tab:inference-time-no-accel}.

\begin{table}[tb]
\centering
\caption{Inference speed per diffusion step for \nameofmethod{} pipeline without engineering-level acceleration.}
\vspace{3pt}
\label{tab:inference-time-no-accel}
\begin{tabular}{ccccc}
\toprule
Task Type & Resolution & Total Frames & Sparse Attention & Time/Step (s) \\
\midrule
T2V & 480p (480 $\times$ 848) & 121  & \ding{55}  & $0.9064 \pm 0.0062$ \\
T2V & 480p (480 $\times$ 848) & 241 & \ding{55} & $1.7015 \pm 0.0203$ \\
\midrule
T2V & 720p (720 $\times$ 1280) & 121 & \ding{55}  & $2.0084 \pm 0.0229$ \\
T2V & 720p (720 $\times$ 1280) & 121 & \checkmark &  $1.5638 \pm 0.0150$ \\
T2V & 720p (720 $\times$ 1280) & 241 & \ding{55} & $5.5070 \pm 0.0284$ \\
T2V & 720p (720 $\times$ 1280) & 241 & \checkmark &  $2.9475 \pm 0.0206$ \\
\bottomrule
\end{tabular}
\end{table}

\paragraph{Inference Speed with Engineering Acceleration.} 
We also evaluate inference speed with basic engineering-level acceleration techniques enabled to demonstrate practical performance achievable in real-world deployment scenarios.
Please note that in this experiment, we do not pursue the most extreme acceleration at the cost of generation quality, but rather to achieve notable speed improvements while maintaining nearly identical output quality.
Several acceleration techniques are employed in this experiment: SageAttention~\cite{zhang2024sageattention2} is adopted for attention operations to reduce memory complexity and improve computational efficiency; the DiT is compiled using PyTorch's \texttt{torch.compile} to enable kernel fusion and operator optimization; a feature caching mechanism is implemented during diffusion sampling to reuse cached intermediate features for non-critical steps, reducing redundant computations.
To better reflect practical performance, we report the total inference time for 50 diffusion steps.
The results are presented in Table \ref{tab:inference-time-accel}.
It can be observed that even with engineering acceleration techniques enabled, sparse attention further reduces the inference time for long-context sequences, demonstrating its effectiveness in complementing engineering optimizations.

\begin{table}[tb]
\centering
\caption{Total inference time for 50 diffusion steps for \nameofmethod{} with acceleration techniques enabled.}
\vspace{3pt}
\label{tab:inference-time-accel}
\resizebox{\linewidth}{!}{
\begin{tabular}{cccccc}
\toprule
Task Type & Resolution & Total Frames & Sparse Attention & Time  & Avg Time/Step   \\
\midrule
T2V & 480p (480 $\times$ 848) & 121  & \ding{55}  & 13.90 & 0.2781 \\
T2V & 480p (480 $\times$ 848) & 241 & \ding{55} & 27.08 & 0.5418 \\
\midrule
T2V & 720p (720 $\times$ 1280) & 121 & \ding{55}  & 28.33 & 0.5667 \\
T2V & 720p (720 $\times$ 1280) & 121 & \checkmark & 26.41 & 0.5283 \\
T2V & 720p (720 $\times$ 1280) & 241 & \ding{55} & 96.78 & 1.9356 \\
T2V & 720p (720 $\times$ 1280) & 241 & \checkmark & 58.39 & 1.1679 \\
\bottomrule
\end{tabular}
}
\end{table}

\paragraph{GPU Memory Requirements.} The GPU memory requirements are contingent upon the hardware configuration and the employment of offloading techniques.
All memory measurements reported herein are conducted on a single GPU configuration.
When pipeline offloading, group offloading, and VAE tiling are enabled, the entire pipeline can complete end-to-end inference for 720p 121-frame T2V/I2V generation with a peak memory of 13.6 GB, thereby enabling inference on a single consumer-grade GPU (e.g., RTX 4090).  

%% file: inference/inference.tex

%% file: conclusion/conclusion.tex
\section{Conclusion}
In this report, we present HunyuanVideo 1.5, a compact 8.3B-parameter Diffusion Transformer (DiT) for open-source video generation that competes with leading proprietary systems. Built via progressive training from a text-to-image model, it achieves high-quality text-to-video and image-to-video synthesis through meticulous data curation and a multi-stage pipeline. The efficient DiT framework integrates SSTA for accelerated inference and a dedicated video super-resolution network for high-resolution output, ensuring state-of-the-art visual quality, motion coherence, and strong multimodal alignment. HunyuanVideo 1.5 exhibits exceptional bilingual prompt understanding, accurate text rendering, and reliable instruction-following. Comprehensive evaluations confirm it sets a new standard among open-source video generators. By releasing this high-performance, lightweight model, we provide an accessible foundation to lower barriers for creative and research applications, fostering broader exploration and innovation.

%% file: contributers/contributer.tex
\clearpage
\section{Project Contributors}

\begin{itemize}
  \item \textbf{Project Sponsors:} Jie Jiang, Linus, Yuhong Liu, Peng Chen 
    \item \textbf{Project Leader:} Zhao Zhong
    \item \textbf{Core Contributors:} (Authors in bolds are project leaders, others are listed alphabetically)
        \begin{itemize}
            \item \textbf{Captioner \& Data:} \textbf{Xin Li, Bing Wu}, Duojun Huang, Hao Tan, Xinchi Deng, Xuefei Zhe, Zhenyu Wang
            \item \textbf{VAE \& Model Accleration:} \textbf{Songtao Liu}, Changlin Li, Chang Zou, Fang Yang, Jianbing Wu,  Jack Peng, Patrol, Peizhen Zhang, Penghao Zhao, Weiyan Wang, Xiao He, Yang Li, Yuanbo Peng
            \item \textbf{Pretraining \& Post Training:} \textbf{Yue Wu}, Jiangfeng Xiong, Qi Tian, Weijie Kong, Yanxin Long, Zuozhuo Dai
        \end{itemize}
    \item \textbf{Contributors (Listed alphabetically):} Bo Peng, Coopers Li, Gu Gong, Guojian Xiao, Jiahe Tian, Jiaxin Lin, Jie Liu, Jihong Zhang, Jiesong Lian, Kaihang Pan, Lei Wang, Lin Niu, Mingtao Chen, Mingyang Chen, Mingzhe Zheng, Miles Yang, Qiangqiang Hu, Qi Yang, Qiuyong Xiao, Runzhou Wu, Ryan Xu, Rui Yuan, Shanshan Sang, Shisheng Huang, Siruis Gong, Shuo Huang, Weiting Guo, Xiang Yuan, Xiaojia Chen, Xiawei Hu, Wenzhi Sun, Xiele Wu, Xianshun Ren, Xiaoyan Yuan, Xiaoyue Mi, Yepeng Zhang, Yifu Sun, Yiting Lu, Yitong Li, You Huang, Yu Tang, Yixuan Li, Yuhang Deng, Yuan Zhou, Zhichao Hu, Zhiguang Liu, Zhihe Yang, Zilin Yang, Zhenzhi Lu, Zixiang Zhou
\end{itemize}